# DTSGAN: Learning Dynamic Textures via Spatiotemporal Generative Adversarial Network

Xiangtian Li[1,a], Xiaobo Wang[2,b], Zhen Qi[3,c], Han Cao[4,d], Zhaoyang Zhang[5,e], Ao Xiang[6,f]

[1]*University of California San Diego, Electrical and Computer Engineering, San Diego, USA*
[2]*Georgia Institute of Technology, Computer Science, Atlanta, USA*
[3]*Northeastern University, Electrical and Computer Engineering, Boston, USA*
[4]*University of California San Diego, Computer Science, San Diego, USA*
[5]*University of California San Diego, Computational Science, San Diego, USA*
[6]*Northern Arizona University, Information Security and Assurance, Arizona, USA*

[a]il160@ucsd.edu, [b]xwang468@gatech.edu, [c]garyzhen79@gmail.com, [d]h2cao@ucsd.edu, [e]zhz088@ucsd.edu, [f]ax36@nau.edu

***Abstract:****Dynamic texture synthesis aims to generate sequences that are visually similar to a reference video texture and exhibit specific stationary properties in time. In this paper, we introduce a spatiotemporal generative adversarial network (DTSGAN) that can learn from a single dynamic texture by capturing its motion and content distribution. With the pipeline of DTSGAN, a new video sequence is generated from the coarsest scale to the finest one. To avoid mode collapse, we propose a novel strategy for data updates that helps improve the diversity of generated results. Qualitative and quantitative experiments show that our model is able to generate high quality dynamic textures and natural motion.*

## 1. Introduction

Consider a dynamic scene in our mind, such as clouds, waves, and geysers. How do we visualize the contents of dynamic videos? We may construct a general structure of a texture image, portray the objects on the image, fill the picture with different colors, and adjust the lightness to address the main object. In addition, we need to transit from the first frame reasonably to generate a realistic dynamic scene. However, in reality, the process of generating a video sequence is far from this simple. Different operations are not independent, and they may occur at the same time. For example, we may simultaneously portray both shapes and motions of the objects, and the color and lightness may change over time. Synthesizing a dynamic texture is an intricate process in space and time.

Learning dynamic textures for video synthesis has been widely explored with both non-parametric and parametric method. For the non-parametric approach, pixels or patches of the given dynamic texture are sampled to synthesize dynamic textures [1,2]. On the other hand, parametric models learn dynamic texture in two manners, statistical methods and deep neural networks. Statistical approaches [3–5] explore the underlying models of the dynamic texture exemplar and synthesize new texture videos by sampling from the learned model. Recently, parametric models have been designed as deep neural networks [6–8]. Although these methods show significant improvements over statistical parametric models, they are still limited in generating dynamic textures with non-local structures.

In this work, we aim to learn dynamic textures via spatiotemporal generative networks (DTSGAN), as shown in Figure 1. We address to construct non-local structures by building a pyramid architecture. DTSGAN learns from a single videoclip from the coarsest scale to the finest one. At each scale, a video generative network maps a random vector to a dynamic texture video. Except for the coarsest scale, the random vector at each scale consists of two parts: the first is a randomly initialized noise, and the second is upsampled from the generated video in the previous scale. To avoid mode collapse, we propose a novel data update strategy that changes the training data every fixed training step sequentially. Once trained, DTSGAN can synthesize diverse high quality dynamic textures and plausible motions from noise. We qualitatively and quantitatively demonstrate that our approach performs favorably against existing methods.

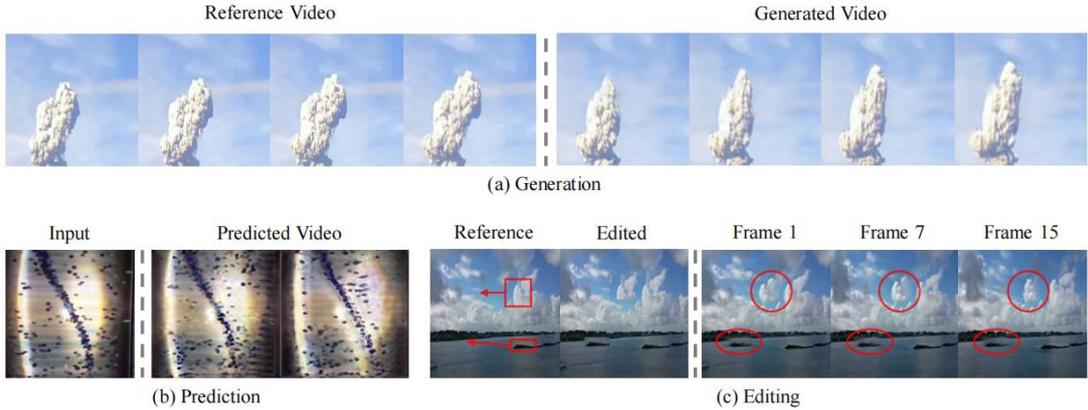

*Figure 1. Overview of our work. We propose to learn dynamic textures from raw videos. The acquired models show the applications of:(a) dynamic textures generation for cloud movement; (b) future prediction for rotary motions; (c) editing dynamic textures from coarsely copy-and-paste inputs to seamless videos.*

We explore the application of our model on future pre- dictions. Given a starting frame, we design an encoder upon Video GAN [9], attach it to the generator at the finest scale, and feed the input texture image into the encoder. Also, we add a loss term to minimize the distance between the input frame and the first frame in the generated video. [10] Extensive experimental results show that our model can generate realistic videoclips conditioned to the input frame.

We make the following five contributions in this work:

• We propose a spatiotemporal generative network that learns dynamic textures from a single videoclip.

• We introduce a novel data update strategy that helps improve the diversity of generated results.

• We demonstrate that the proposed algorithm performs favorably against the state-of-the-art methods through extensive experiments.

• We design an encoder attached to the network for future predictions.

• We show the proposed method can be used for video editing based on reference patches.

## 2. Related Work

Dynamic texture synthesis. Dynamic textures are sequences of images of moving scenes that exhibit certain stationary properties in time; this includes sea-waves, smoke, foliage, whirlwind but also talking faces, and traffic scenes, to name a few [3]. Dynamic texture synthesis is a process of creating artificial textures, which plays a vital role in video analysis and has been widely explored for years. [11]

There are two main approaches to synthesize dynamic texture: non-parametric synthesis and parametric models. A non-parametric synthesis is a sampling approach that synthesizes dynamic textures by sampling pixels or patches of a given source dynamic texture. [12] Kwatra et al. [1] em- ploy graph cut for dynamic texture synthesis to achieve high visual quality results. Lizarraga-Morales et al. [2] extend the texture synthesis method to dynamic texture synthesis. It explores the use of local spatiotemporal features to capture the structure in both spatial and temporal domains and describes the motion and appearance of dynamic textures. A conditional generative CNN (cgCNN) [13] defines an energy-based conditional distribution of a given dynamic texture exemplar. New dynamic textures can be synthesized from the learned conditional distribution. However, non- parametric methods always fail to synthesize dynamic textures without spatial stationary and preserve temporal continuity of the synthesized dynamic texture sequences. [14]

Parametric methods estimate the latent space of dynamic textures. Early methods focus on statistical models, such as linear dynamic system (LDS) [15,3], and closed-loop LDS (CLDS) [5]. Recently spatiotemporal Convolutional Neural Networks (CNNs) are proposed to generate samples of dynamic textures. Funke et al. [6] propose a Bayesian- based nonlinear dynamic texture modeling method for dynamic texture synthesis. It utilizes the Gaussian process la- tent variable model for dimensional reduction and designs a multi-kernel dynamic system for the latent dynamic behavior modeling [16]. In the follow-up work [8], Funke et al. introduce a parametric model for dynamic textures based on the feature representations of a CNN trained on object recognition [4]. In [7], a two-stream CNN, one for static texture appearance of each frame and the other for temporal variations between frames, is proposed for dynamic texture synthesis [17]. Xie et al. [18] propose a dynamic generator (DG) model to execute a nonlinear transformation from a la- tent space vector to each timeframe of the video sequence. The limitation of current CNN methods is that they can- not synthesize dynamic textures with non-local structures because the optimizations are always trapped in local mini- mum during training [19].

Single-image generative adversarial networks. Recently, single-image generative models have been proposed to capture the internal distribution of a single image. [20] InGAN [21] learns the input image's internal patch-distribution and generates images of different sizes with the same internal patch-distribution as the input image. Multi-scales methods such as SinGAN [22] also have been proposed to learn on a single image and generate high quality, diverse samples with the same patch distribution as the input image. [23] The seminal work of SinGAN can also be used for conditional image generation tasks [24,25]. Tu- iGAN [24] aims to capture the distribution variations be- tween two unpaired images and transform an image from one source distribution to the other. Vinker et al. [25] pro- pose a single-image based method to train deep generative models that can perform challenging image manipulation tasks. However, these models all focus on image generation and manipulation and do not pay much attention to video generation. [26] In this work, we propose a model that em- ploys the single-data based strategy to generate videos by 3D convolution. Instead of learning the patch distribution on the timeaxis [27], we use a data update strategy to accommodate GPUs' limited memory capacity [28].

Video generation. Video generation aims to map a random noise to the spatial content and temporal motion, which form a realistic video sequence. VideoVAE [29] proposes a cross convolutional network to encode the image and motion information as feature maps and learned kernels, respectively. The network can recurrently synthesize possible future frames of the input image. VideoGAN [9] utilizes generative adversarial networks (GANs) to synthesize videos. Combining the static background stream and the dynamic foreground stream, the model can generate realis- tic short videos. [30] Instead of generating foreground and background separately, TGAN [31] maps a latent vector to a set of latent variables through a temporal generator and then transforms the latent variables into a video sequence. [32] MoCoGAN [33] can learn to disentangle motion from content by sampling from separate latent spaces and generate a videoclip by recurrently generating video frames. Built upon the state-of-the-art BigGAN architecture [34], DVD- GAN [35] is capable of capturing the complexity of a large video dataset as well as producing longer and higher-quality video samples.

## 3. Proposed Method

### 3.1. Overview

In this section, we present a generative model trained on a single video. We briefly discuss the SinGAN [22] method before describing the details of DTSGAN. The SinGAN model consists of a pyramid of generators $\{G_0,..., G_N\}$. These generators are sequentially trained on videos $\{x_0,..., x_N\}$ from the coarsest scale to the finest one. Before being fed into a sequence of convolution layers of $G_n$ ($n > 0$), the noise $z_n$ is added to the image $(\tilde{x}_{n-1})\uparrow^r$ that is an upsampled version of the image $x_{n-1}$ from the coarser scale. This ensures that the GAN can generate the missing details in $(\tilde{x}_{n-1})\uparrow^r$ and also does not disregard the noise.

We use a multi-scale architecture similar to that of the SinGAN. However, unlike SinGAN that can only generate 2D images, our model is able to synthesize a videoclip with the same patch distribution as the input video sequence.

### 3.2. Model Architecture

Multi-scale architecture. Similar to SinGAN, we employ the multiscale architecture in our work. The architecture of our model is shown in Figure 2. The generators and discriminators at each scale have the same architecture.

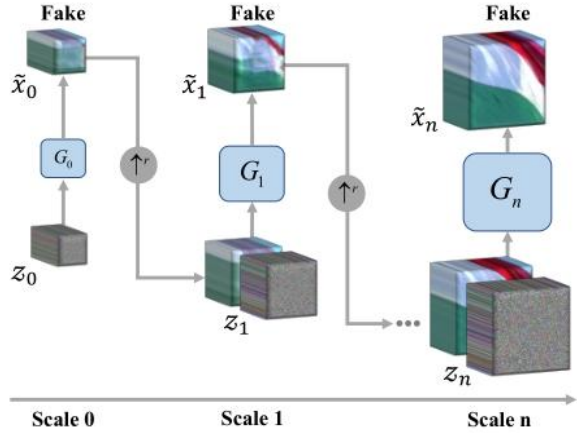

*Figure 2. Architecture overview. Our model uses a multi-scale architecture, where both the generator and discriminator progress in a coarse-to-fine manner. Starting from scale 0, the generator $G_n$ takes in the summation of the upsampled video $(\tilde{x}_n) \uparrow^r$ and a random noise $z_n$, while the input of the generator at scale 0 is a random noise. At each scale, we update the videoclip data every fixed step to avoid mode collapse during training.*

Since the input is a 16-frame video, all the operations in our model are in a 3D manner, i.e., 3D convolution, 3D deconvolution, and 3D batch normalization. At the coarsest scale, $G_0$ maps the 3D Gaussian noise $z_0$ to an 3D video $\tilde{x}_0$ by

$$\tilde{x}_0 = G_0(z_0). \quad (1)$$

An upsampling process, enlarging the size while main-taining the length of the video, is then applied to $\tilde{x}_0$. The following generators $G_n$ at finer scales ($n > 0$) take in the noise $Z_n$ and the upsampled version of the video from the coarser scale, i.e.,

$$\tilde{x}_n = G_n(z_n, (\tilde{x}_{n-1}) \uparrow^r), \quad n > 0. \quad (2)$$

At the finer scales, the model captures the patch distribution of the input video with a relatively small receptive field, which generates details of the video sequence [28].

**Generator.** The generator is a 3D model that takes in a 3D video input. We stack 16 consecutive frames from a single video together to create an input. Except for the last convolution layer, each convolution layer is followed by a 3D batch normalization [36] and a leaky ReLU layer. A hyperbolic tangent activation function is used at the last convolution layer. The size of the output feature map is the same as the input.

**Discriminator.** The discriminator distinguishes whether the generated video is real. The patch discriminator has a receptive field of $11 \times 11 \times 11$. Each convolution layer is followed by a 3D batch normalization and a leaky ReLU layer except the last convolution layer. [37] The last convolution layer directly outputs a 3D feature map with only one channel. Then we use the mean of the output feature map to compute the adversarial loss.

### 3.3. Data Update Strategy

A video generative model needs to learn the motions in videos over time. With a receptive field of $11 \times 11 \times 11$, the generators and discriminators can see a small patch in the height and width axis while a relatively large patch in the timeaxis. Therefore, a model trained on only one 16-frame videoclip fails to generate results with high diversity.

In order to improve the diversity of the generated results, the model needs to see more motions over time. One possible solution is increasing the length of a training video clip. However, this method fails in practice because of limited computational capacity. The 3D model entails a large amount of space that causes the training process to collapse. Another approach is the data update strategy, i.e., feeding different videoclips to the model every fixed training step. A direct idea is randomly selecting a video clip from a single video every fixed step. However, since the video clips vary significantly in shapes and motions over time, this model fails to capture the key information of a video and generates chaotic results.

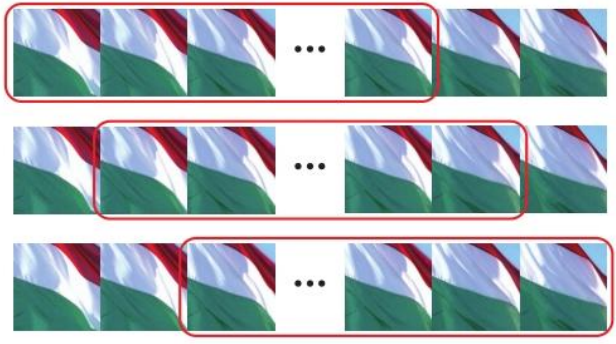

*Figure 3. Data Update Strategy. We update the training video clip every fixed training step to make the model see more motions over time.*

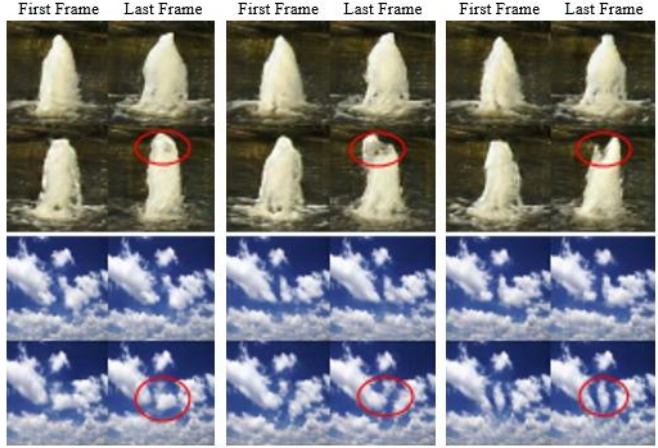

*Figure 4. Data update strategy comparison. We present two examples here. For each example, the first row is the model training on a single videoclip, and the second row is the model employing the data update strategy.*

Instead of randomly updating the training data, we propose a new data strategy, as shown in Figure 3. During training, we update the previous videoclip with its neighbor one. As the training progresses, the model can see different parts of the input video from a single video and generate videos with higher diversity in motions and textures. Since the neighbor videoclips are similar in motions and textures, the model can learn the patch distribution of dynamic textures more steadily. The qualitative comparison is shown in Figure 4.

### 3.4. Model Training

Adversarial loss. The adversarial loss builds on the fact that the generator generates realistic videos to fool the discriminator, and the discriminator distinguishes the real videos from generated ones. At each scale n, there are a generator $G_n$ and a discriminator $D_n$. We use the WGAN-GP loss [38] as the adversarial loss, which can make the training process more steady:

$$L_{D_n} = D_n(x_n) - D_n(G_n(\tilde{x}_n)) - \lambda(\|\nabla_{\hat{x}_n} D_n(\hat{x}_n)\|_2 - 1)^2. \quad (3)$$

where, $\hat{x}_n = \alpha x_n + (1-\alpha)\tilde{x}_n$, $\alpha \sim (0,1)$, $\lambda$ is a given penalty coefficient.

Reconstruction loss. We add the reconstruction loss to ensure that the model can generate the input videoclip. We set $z_0$ as a fixed noise map and $z_n$ as zero for all $n > 0$ to reconstruct the input videoclip. The reconstruction loss is

$$L_{rec} = \|\tilde{x}_n^{rec} - x_n\|^2 \quad (4)$$

where, $\tilde{x}_n^{rec}$ is the generated video at scale n.

Loss function. The overall loss is the sum of the adversarial loss and the reconstruction loss, weighted by a hyper-parameter η, which is

$$L = L_{D_n} + \eta L_{rec}. \tag{5}$$

## 4. Experiments and Analysis

We conduct experiments on a large variety of videos, both qualitatively and quantitatively.

### 4.1. Experimental Settings

The videos come from the Dynamic Texture DataBase (DTDB) [39]. We set the coarsest scale as 25px and the finest scale as 150px. The scaling factor r is approximately 1.39, and the number of scales is 8. As for the data update frequency, we find that 50-200 is a fit range of epochs that increases the diversity of generated results and maintains the quality of the videoclips.

We compare our model with three baseline methods, which are Two-Stream [7], DG [18], and cgCNN [13]. Two- Stream model utilizes an appearance stream and a dynamics stream to synthesize dynamic textures. DG is a dynamic generator learned by an alternating back-propagation algorithm through time. cgCNN is an exemplar-based model for dynamic texture synthesis. We select one example from each category (18 in total) from the test set of the DTDB for quantitative evaluation. We compare the results from these baseline methods and our model, both qualitatively and quantitatively.

### 4.2. Qualitative Evaluation

We evaluate the proposed DTSGAN model on several videos from the test set of DTDB [39]. Some generated videoclips are shown in Figure 5. Our model can generate a 16-frame videoclip with a size of 150 × 150 on a single GPU. More generated videos can be found in the supple- mentary material.

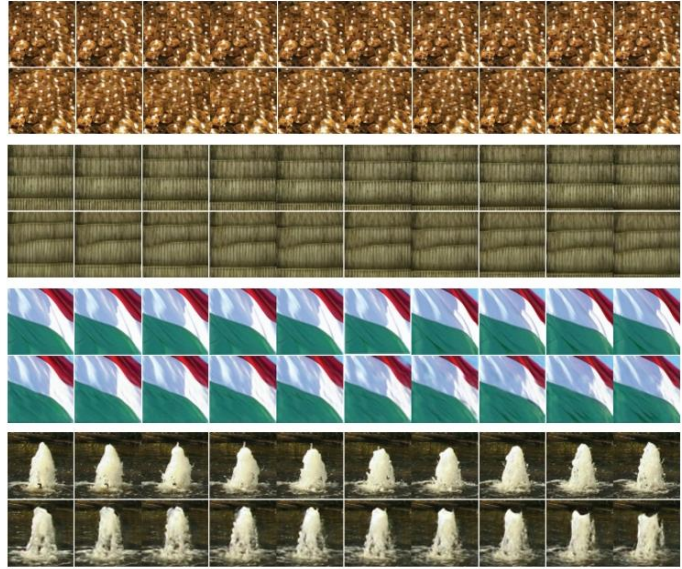

*Figure 5. Qualitative results. After training DTSGAN on a single video, our model can generate a 16-frame realistic video with a size of 150 × 150. For each pair, the first row is the original videoclip and the second is the generated video. We display the first eight frames for each example.*

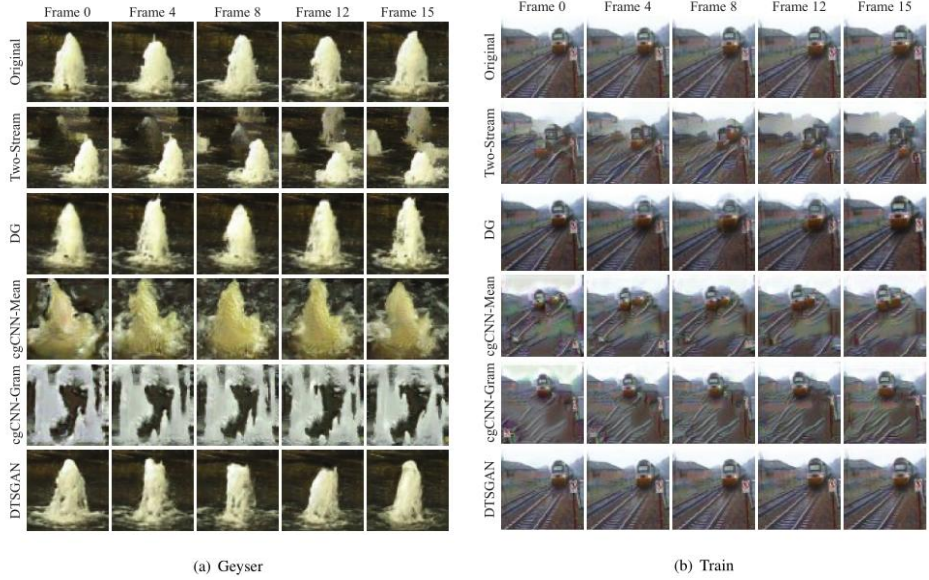

(a) Geyser  (b) Train

*Figure 6 presents the qualitative comparison between DTSGAN and three baseline methods. We notice that our results exhibit obvious advantages over other methods in maintaining shapes, textures, and motions.*

### 4.3. Quantitative Evaluation

Multi-scale structural similarity. We measure the multi- scale structural similarity (MS-SSIM) [40] between the original video and the generated video. A higher MS-SSIM score indicates higher visual realism of generated results. Given two videoclips, we compute the MS-SSIM score be- tween each frame in the original videos and the corresponding frame in generated videos. The average of MS-SSIM metric represents the visual realism of the generated video.

Frechet Inception Distance. The Frechet Inception Distance (FID) metric [41] measures the distance between the feature distribution of original and generated images. We use the FIN metric to assess the realism of generated video clips. For a generated video clip $X = \{\tilde{x}_1, \tilde{x}_2, \cdots, \tilde{x}_T\}$, we compute the FID between the frames in the generated video clip and the frames in the reference video clip $X = \{x_1, x_2, \cdots, x_T\}$. In particular, we set T = 16 and train our model on 16-frame videoclips to generate videos.

Standard deviation of the normalized LPIPS. The standard deviation of the normalized LPIPS (δ-N-LPIPS) [42] can measure the smoothness of a video. Given a generated video $\tilde{x}_n$, we compute the smoothness between frame video $\tilde{x}_n$, we compute the smoothness between frame $\tilde{x}_n^i$ and frame $x_n^{i+1}$ as

$$\sigma\left(\frac{\text{LPIPS}(\tilde{x}_n^i, \tilde{x}_n^{i+1})|i = 1, ..., T-1}{\text{LPIPS}(\tilde{x}_n^0, \tilde{x}_n^T)}\right), \quad (6)$$

where, σ stands for the standard deviation. Lower δ-N- LPIPS implies a smoother transition between frames. In the experiment, we compute all the ratios of the LPIPS scores between every two consecutive frames and the LPIPS score between the first frame and the last frame.

*Table 1. Quantitative comparisons on selected videos from the DTDB. We compute the value of MS-SSIM, FID, and δ-N-LPIPS on the videos generated by the five models separately. We use the average score of the 18 videos to measure the models. The best results under different metrics are marked by Bold.*

| Method | MS-SSIM ↑ | FID ↓ | δ-N-LPIPS ↓ |
|---|---|---|---|
| Two-Stream | 0.150 | 229.464 | 0.471 |
| DG | 0.609 | 223.232 | 0.768 |
| cgCNN-Gram | 0.217 | 240.834 | 0.864 |
| cgCNN-Mean | 0.231 | 255.231 | 0.912 |
| DTSGAN (Ours) | **0.621** | **193.063** | **0.223** |

Analysis. Figure 6 and Table 1 show qualitative and quantitative comparisons of these methods. As can be seen in Figure 6, Two-Stream and cgCNN do not generate results similar to the reference video. Thus, they have a low MS-SSIM score and high FID value, as shown in Table 1. Although DG generates relatively similar results to the original video in texture distribution, the generated videos contain limited smoothness that achieves a high δ-N-LPIPS score. In contrast, our model leads to the highest value of MS-SSIM and lowest value of FID and δ-N-LPIPS among all the experiment methods, which demonstrates the proposed model can generate high quality and smooth transition videos.

Table 2. User study results. We conduct a user study to ask participants to choose their favorite videoclips. Each number represents the average preference percentage of each method.

| Method | Percentage |
|---|---|
| Two-Stream | 11.78% |
| DG | 20.67% |
| cgCNN-Gram | 5.89% |
| cgCNN-Mean | 7.78% |
| **DTSGAN (Ours)** | **53.89%** |

User study. We conduct a user study to compare the quality of our generated video clips with those product from the other four methods. We utilize the 18 dynamic texture videos in the previous section and use these five methods to generate video clips. For each dynamic texture video, we show five results generated by the methods separately to participants and ask them to select the most realistic one. Fifty people participate in the study. Table 2 shows that the majority of participants prefer the results generated by our DTSGAN model compared with other results.

*4.4. Ablation Study*

In this section, we evaluate the effect of data update strategy. We use LPIPS [43] to measure the diversity of synthesized results. We train two models for each reference video, one learning from several video clips with training data updating, the other only learning from one videoclip. Given two videos generated by the same model, we compute the LPIPS between each frame in one video and the corresponding frame of the other video. We set the average of LPIPS scores of each frame pair as the LPIPS between two videos. For each video category, we calculate the difference between every two videos among 20 videos. The mean of the LPIPS scores among all video pairs represents the diversity of the results in one category. [44]

Table 3. Effect of the data update strategy. We conduct an ablation study on the data update strategy. Each number represents the average LPIPS among the 20 synthesized videos of each video category. A higher LPIPS score represents better diversity.

| Category | With data update | w/o data update |
|---|---|---|
| Geyser | 0.312 | 0.236 |
| Escalator | 0.177 | 0.103 |
| Candy | 0.416 | 0.368 |
| Coral | 0.143 | 0.091 |
| Flag | 0.179 | 0.146 |

The quantitative results are shown in Table 3. From these results, we can find that the data update strategy improves the diversity of the generated results remarkably.

**5. Conclusions**

In this paper, we introduce DTSGAN, a model designed to capture the spatial and temporal distribution of dynamic textures for accurate and diversified video generation. We build the 3D model to learn dynamic textures from coarse scales to fine ones, and utilize sequential training data update with fixed steps to improve the diversity of generated clips. Experimental results demonstrate that DTSGAN outperforms Two-Stream, DG, and cgCNN both qualitatively and quantitatively. However, a limitation of our method, along with other single-video approaches, is that we need to train a model from scratch for every dynamic texture video to synthesize new videos.